\pdfoutput=1

\documentclass[11pt]{article}

\usepackage[]{acl}

\usepackage{times}
\usepackage{latexsym}

\usepackage[T1]{fontenc}

\usepackage[utf8]{inputenc}

\usepackage{microtype}

\usepackage{inconsolata}

\usepackage[utf8]{inputenc} 
\usepackage[T1]{fontenc}    
\usepackage{hyperref}       
\usepackage{url}            
\usepackage{booktabs}       
\usepackage{amsfonts}       
\usepackage{nicefrac}       
\usepackage{microtype}      
\usepackage{color,xcolor}         

\usepackage{inconsolata}
\usepackage{enumitem}
\setenumerate[1]{itemsep=0pt,partopsep=0pt,parsep=\parskip,topsep=5pt}
\setitemize[1]{itemsep=0pt,partopsep=0pt,parsep=\parskip,topsep=5pt}
\setdescription{itemsep=0pt,partopsep=0pt,parsep=\parskip,topsep=5pt}

\usepackage{amsmath}
\usepackage{nicefrac}   
\usepackage{wrapfig}
\usepackage{setspace}
\usepackage{makecell}
\usepackage{multirow}
\usepackage{amssymb}
\usepackage{calc}
\usepackage{graphicx}
\usepackage{subfigure}
\usepackage{physics,amsmath}

\definecolor{grass}{rgb}{0.18,0.80,0.18}

\newcommand{\Model}{\textsc{LoGiPT}}

\usepackage[customcolors]{hf-tikz} 
\usepackage{soul}

\definecolor{lightyellow}{rgb}{1.0, 1.0, 0.6} 
\definecolor{lightcyan}{rgb}{0.48, 1.0, 1.0}
\definecolor{lightblue}{RGB}{173, 216, 230}
\definecolor{lightpink}{RGB}{255, 182, 193}
\definecolor{lightgreen}{RGB}{144, 238, 144}
\definecolor{lightlavender}{RGB}{230, 230, 250}
\definecolor{lightpeach}{RGB}{255, 218, 185}
\definecolor{lightaqua}{RGB}{173, 216, 230}
\definecolor{lightmint}{RGB}{152, 255, 152}
\definecolor{lightcoral}{RGB}{240, 128, 128}
\definecolor{lightgray}{RGB}{211, 211, 211}
\definecolor{lightrose}{RGB}{255, 182, 193}
\definecolor{lightturquoise}{RGB}{173, 216, 230}
\definecolor{lightbeige}{RGB}{245, 245, 220}
\definecolor{lightorchid}{RGB}{230, 168, 215}
\definecolor{lightgold}{RGB}{255, 223, 186}
\definecolor{lightgray}{RGB}{211, 211, 211}

\usepackage{tcolorbox}

\colorlet{mythmback}{lightcyan!40!white}
\newtcolorbox{boxEnv}{
center,
width=0.95\linewidth,
boxrule=0.5pt,
left=10pt,right=10pt,
top=2pt,bottom=2pt,
before skip=10pt, after skip=10pt, 
}

\newtcolorbox{caseBoxEnv}[2][]{
center,
width=\linewidth,
boxrule=0.5pt,
left=2pt,right=2pt,
top=2pt,bottom=2pt,
fonttitle=\bfseries,
title=#2,#1,
colbacktitle=gray!60!black 
}

\title{Language Models can be Logical Solvers}

\author{Jiazhan Feng$^1$\thanks{\quad Work done during Jiazhan's internship at Microsoft Azure AI.} \quad Ruochen Xu$^2$ \quad Junheng Hao$^2$  \quad Hiteshi Sharma$^2$\\ \quad {\bf Yelong Shen}$^2$ \quad {\bf Dongyan Zhao}$^1$ \quad {\bf Weizhu Chen}$^2$  \\
  $^1$Peking University, Beijing \quad $^2$Microsoft Azure AI, Redmond \\
  \texttt{\{fengjiazhan,zhaody\}@pku.edu.cn} \\
  \texttt{\{ruox,junhenghao,hitshar,yeshe,wzchen\}@microsoft.com} \\
  }

\begin{document}
\maketitle
\begin{abstract}
Logical reasoning is a fundamental aspect of human intelligence and a key component of tasks like problem-solving and decision-making. Recent advancements have enabled Large Language Models (LLMs) to potentially exhibit reasoning capabilities, but complex logical reasoning remains a challenge. The state-of-the-art, solver-augmented language models, use LLMs to parse natural language logical questions into symbolic representations first and then adopt external logical solvers to take in the symbolic representations and output the answers. Despite their impressive performance, any parsing errors will inevitably result in the failure of the execution of the external logical solver and no answer to the logical questions. In this paper, we introduce \Model, a novel language model that directly emulates the reasoning processes of logical solvers and bypasses the parsing errors by learning to strict adherence to solver syntax and grammar. \Model~is fine-tuned on a newly constructed instruction-tuning dataset derived from revealing and refining the invisible reasoning process of deductive solvers. Experimental results on two public deductive reasoning datasets demonstrate that \Model~outperforms state-of-the-art solver-augmented LMs and few-shot prompting methods on competitive LLMs like ChatGPT or GPT-4.
\end{abstract}

\section{Introduction}

Logical reasoning is a foundational element of human intelligence, holding a pivotal role in tasks like problem-solving, decision-making, and critical thinking~\citep{huang-chang-2023-towards}. Recently, substantial advancements have been achieved in the field of NLP through the development of large language models (LLMs)~\citep{chatgpt2022,openai2023gpt4,bard2023,touvron2023llama,touvron2023llama2}. It has been noted that language models (LMs) could potentially display reasoning capabilities when they reach a certain scale threshold (e.g., training compute, model parameters, etc.)~\citep{kaplan2020scaling,wei2022emergent,hoffmann2022training}. To this end, LLMs can answer logical questions with explicit reasoning steps when prompted with a simple snippet: ``\textit{Let's think step by step.}''~\citep{kojima2022large} or step-wise explanations of reasoning (i.e., ``chain of thoughts'')~\citep{wei2022chain}.

While LLMs have made significant progress, complex logical reasoning remains challenging~\citep{valmeekam2022large,liu2023evaluating}. Some prior work~\citep{tafjord-etal-2022-entailer,ling2023deductive} aimed to enable LMs to perform logical reasoning via specialized module fine-tuning, where reasoning is in natural language (NL). However, the ambiguity and complexity of NL can lead to undesired issues like hallucinations and unfaithful reasoning~\citep{saparov2023language,gao2023pal}. To this end, recent work has begun to augment LLMs with access to external \textbf{Solvers}~\citep{chen2022program,ye2023satisfiability,pan2023logic}. In this paper, we focus on the logical solvers, which are theorem provers that can be any automated reasoning tool for checking the truth value of logical formulas in symbolic language (SL). Invoking logical solvers can guarantee the accuracy of logical reasoning and relieve the burden of LLMs to execute intricate and precise deductive reasoning.

\begin{figure*}[t!]
  \centering   
  \includegraphics[width=0.9\textwidth]{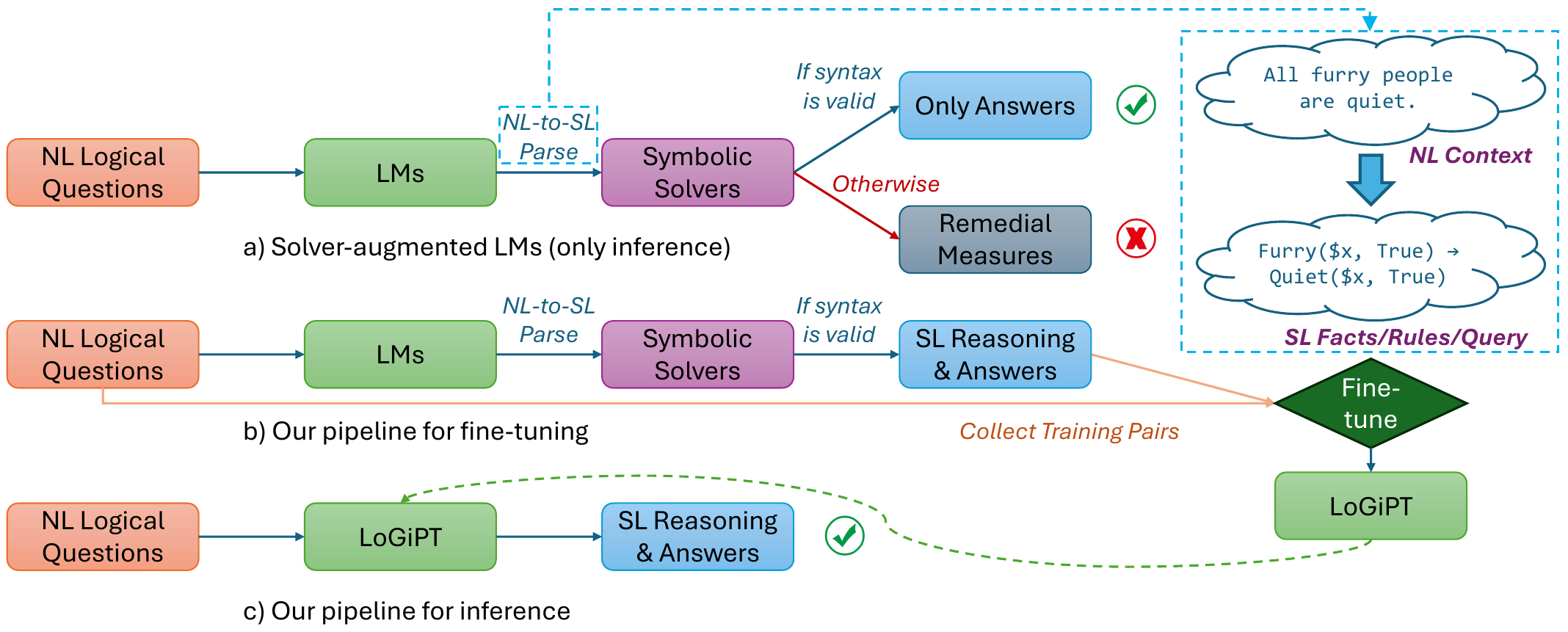}
  \vspace{-2mm}
  \caption{Data flow of current solver-augmented LMs for inference \textbf{(a)}, and our pipeline for \Model~\textbf{(b,c)}.
  }
  \label{fig:dataflow}
  \vspace{-2mm}
\end{figure*}

The data flow of the aforementioned solver-augmented LMs is depicted in Figure~\ref{fig:dataflow}(a). At the outset, the information of logical questions is stored in NL. It is subsequently fed into a LM for parsing into a symbolic representation suitable for solver-input format. Finally, the SL information is dispatched to a symbolic solver, which yields the truth value of the logical question. However, during this process, any NL-to-SL parsing errors will inevitably result in the failure of the reasoning process and no answer to the question. In our preliminary experiments, we observed that the parsing successful rate (i.e., percentage of executable logical formulations) of Vicuna-13B~\citep{vicuna2023} on ProofWriter~\citep{tafjord-etal-2021-proofwriter} is only 17\%, significantly below the expected performance. In addressing parsing failures, current methods either directly use LLMs to reason in NL solely or rely on the solver's erroneous message to regenerate parsing results, but these approaches don't fundamentally resolve the problem.

In this paper, we introduce \Model, a novel LM designed to mimic the reasoning process of logical solvers, enabling it to solve deductive reasoning tasks. We first construct an instruction-tuning dataset containing NL logical questions and their corresponding solver's symbolic reasoning process. After filtering out cases having invalid syntax, we fine-tune open-source LMs like Vicuna or CodeLlama~\citep{roziere2023code} with this data to create \Model. Then, \Model~can generate all implied facts given premises and rules, allowing us to determine the truth value of a logical query by matching it with implied facts or outputting `unknown' if it cannot be determined. The data flow of our pipeline is presented in Figure~\ref{fig:dataflow}(b,c). We can bypass the syntax or grammatical errors derived from NL-to-SL parsing by directly outputting the answers with a fine-tuned \Model. 

Our approach is akin to the process of distillation, whereby we distill knowledge from a symbolic model (i.e., solver) into a neural network (i.e., LM). 
However, the reasoning process of solvers is invisible to users and we can \textbf{only} obtain the answers without intermediate reasoning steps. 
We design a pipeline to reveal and formalize solvers' invisible reasoning processes, creating instruction-tuning datasets with visible and interpretable symbolic reasoning steps (see Figure~\ref{fig:ExampleSimplfied}). 

Our main contributions are three-fold:
\begin{itemize}
    \item To the best of our knowledge, we are the first to propose empowering LLMs to directly learn the reasoning process of logical solvers, thereby acquiring similar reasoning capability for addressing deductive reasoning tasks.

    \item Our proposed \Model, can directly act as a deductive solver and output all Facts implied from NL logical questions while bypassing the syntax or grammatical errors derived from NL-to-SL parsing of solver-augmented LMs.

    
    \item Evaluation results on two public deductive reasoning datasets show that \Model~can outperform state-of-the-art solver-augmented LMs, and few-shot prompting methods on competitive LLMs like ChatGPT or GPT-4.

\end{itemize}

\section{Preliminary}

\subsection{Deductive Reasoning}
\begin{figure}[t!]
  \centering   
  \includegraphics[width=0.45\textwidth]{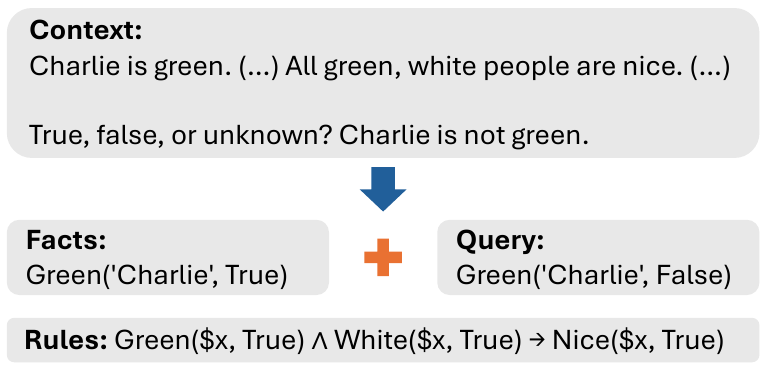}
  \caption{A deductive reasoning question derived from ProofWriter and its parsed Facts, Rules, and Query.}
\vspace{-2mm}
 \label{fig:DeductiveReasoningCase}
\end{figure}

Deductive reasoning is an essential type of logical reasoning problem. It typically commences with known facts and rules from logical context, then proceeds through a series of inference steps until the query can be proved or disproved~\citep{poole2010artificial}. In this paper, we consider the \texttt{Prolog} logic programming language~\citep{clocksin2003programming,korner2022fifty}, which stands as the most prominent symbolic language for describing deductive reasoning problems. 
We showcased a deductive reasoning question along with its corresponding \texttt{Prolog} syntax representation in Figure~\ref{fig:DeductiveReasoningCase}. 

For each question, we denote the NL description as \textbf{Context}. The \textbf{Context} can further be parsed into \textbf{Facts}, \textbf{Rules}, and \textbf{Query}\footnote{In this paper, the term `Query' refers to a specific sentence of statement or comment, while `question' is used in a broader sense to denote the description of a logical problem.}. Specifically, a \textbf{Fact} $F = P(a_1, \cdots, a_t)$ is a symbolic statement with a predicate $P$ and $t$ arguments $\{a_1, \cdots, a_t\}$ where $a_i$ can be a variable, entity, number or bool. For example, \texttt{Green('Charlie', True)} means ``Charlie is green''; \textbf{Rules} are presented in the form of clauses $F_1 \wedge \cdots \wedge F_m \to F_{m+1}  \wedge \cdots \wedge F_n$, where $F_i$ is a Fact. The Rule means ``if each $F_i \in \{F_1,\cdots,F_m\}$ is true, then we can imply that all Facts in $\{F_{m+1},\cdots,F_n\}$ are also true.'' For example, \texttt{Furry(\$x, True) → Quiet(\$x, True)} indicates if variable \texttt{\$x} is furry, then \texttt{\$x} is quiet; a \textbf{Query} $Q$ is also in the format of a Fact that needs to be proved based on Facts and Rules.

\subsection{Solver-augmented LMs}

Solver-augmented LMs have demonstrated remarkable performance in deductive reasoning tasks. As shown in Figure~\ref{fig:dataflow}(a), these model can be generally divided into two stages: \textit{Problem Formulation (from LMs to Symbolic Solvers)} and \textit{Symbolic Reasoning (from Symbolic Solvers to Answers)}.

In \textit{Problem Formulation} stage, an LM is used to parse an NL logical question into symbolic representation (Figure~\ref{fig:DeductiveReasoningCase}). The process can be accomplished by providing LM with detailed instructions about the grammar of \texttt{Prolog}, alongside a few demonstrations as in-context examples~\cite{ouyang2022training}. The LM is expected to identify the symbolic Facts, Rules, and Query from the NL logical question following the instructions; In \textit{Symbolic Reasoning} stage, a solver takes in the symbolic representation obtained in the previous stage and conducts symbolic reasoning. The reasoning process of external off-the-shelf solver, e.g., \texttt{pyke} expert system~\citep{frederiksen2008applying}, is deterministic and invisible. Then, the truth value of the parsed Query, which is the only output of the solver, can be treated as the answer to the given question.  

\subsection{Analysis on the Parsing Successful Rate}

\begin{table}[!t]
\centering
\resizebox{0.45\textwidth}{!}{
    \begin{tabular}{lcc}
    \toprule
        Model & ProofWriter & PrOntoQA \\ 
        \midrule
        Vicuna-13B  & 17.00 & 40.80 \\
        CodeLlama-13B-Base  & 0.33 & 0.40 \\
        CodeLlama-13B-Instruct  & 71.33 & 77.80 \\

\bottomrule
   
    \end{tabular}
} 
    \caption{Parsing successful rate (\%) of our selected open-source LLMs on two deductive reasoning datasets.}
    \label{tbl:parsing_rate}
\end{table}

Through the aforementioned two phases, once the solver-augmented LMs correctly formulate the problem, the answers obtained through symbolic reasoning will be faithful, attributed to the deterministic nature of the solver. However, this heavily relies on the in-context learning capabilities of LMs. 
Therefore, we first calculate the parsing successful rate of three selected open-source LLMs on two deductive reasoning datasets in Table~\ref{tbl:parsing_rate}. Firstly, we observe that CodeLlama-13B-Base (\texttt{CodeLlama-13b-hf}) is unable to effectively conduct NL-to-SL parsing due to the limited in-context learning capabilities in natural languages. Then we can find that replacing the Base model with the Instruct version (\texttt{CodeLlama-13b-Instruct-hf}) can alleviate this issue, which may be attributed to the fact that the Instruct version is further fine-tuned with an additional approx.~5B tokens to better follow human instructions. Overall, open-source LLMs still exhibit parsing performance significantly lower than expected in some cases. 

\begin{figure*}[t!]
  \centering   
  \includegraphics[width=0.95\textwidth]{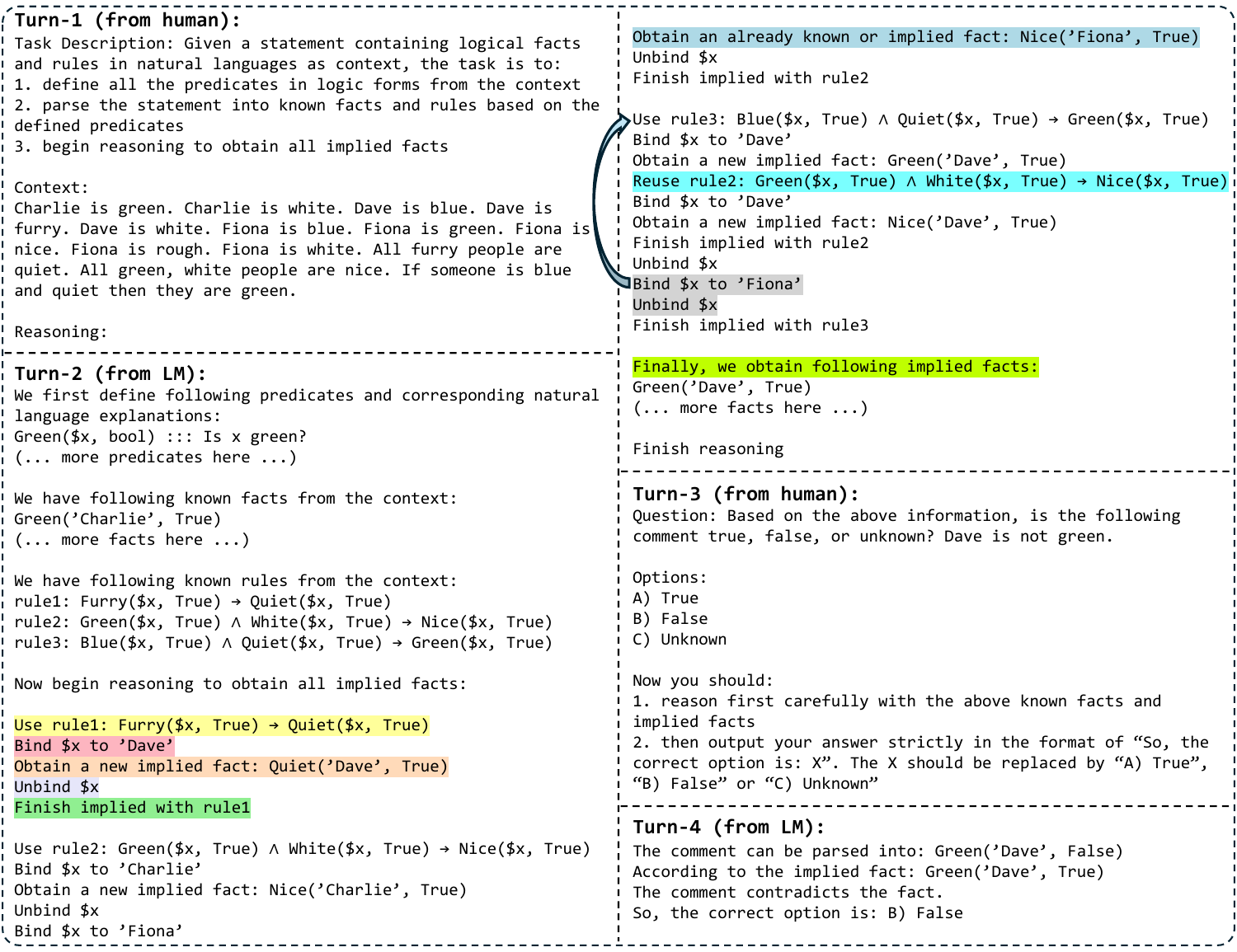}
  \caption{A comprehensive 4-turn training example of our instruction-tuning data. We highlight the initial occurrences of each functionality described in \S\ref{sec:reveal} using the corresponding colors. We omit some predicates and Facts in Turn-2 due to limited space. \textit{Hint: this figure is color-sensitive.}}
 \label{fig:ExampleSimplfied}
\end{figure*}

\section{LoGiPT}

In this paper, we aim to mitigate the parsing issue and present a novel LM, \Model~instructed to imitate the logical reasoning process of Solvers for deductive reasoning tasks. To achieve this, we first reveal the solver reasoning process when solving logical problems (\S\ref{sec:reveal}). Then, we construct a solver-derived instruction-tuning dataset, comprising NL logical questions and corresponding SL reasoning steps (\S\ref{sec:construct}). Finally, we fine-tune open-source LLMs using this dataset to develop \Model~(\S\ref{sec:finetune}).

\subsection{Revealing the Solver Reasoning Process}
\label{sec:reveal}
Before operating on the solvers, we first adopt \texttt{gpt-4} as the problem formulator for NL-to-SL parsing with instructions about the grammar and few-shot demonstrations\footnote{Detailed instructions for NL-to-SL Parsing are shown in Appendix~\ref{apd:instructions_proofwriter} and~\ref{apd:instructions_prontoqa}.}, and obtain the SL representations of all training logical questions of the given logical datasets. Then, consistent with solver-augmented methods, we adopt \texttt{pyke} expert system as the symbolic solver in this work that can make inferences using the \texttt{Prolog} symbolic language. Given a logical question, \texttt{pyke} first sets up a knowledge base and injects all known Facts and Rules (Figure~\ref{fig:DeductiveReasoningCase}) from solver's inputs. Then, it iteratively applies Rules on already known or implied Facts, aiming at obtaining more implied Facts until the Query is proved or disproved.

The reasoning process executed by \texttt{pyke} solver is invisible to users and solver-augmented LMs use the solver as a \textit{black-box}. We hypothesis the `chain-of-thought' reasoning process of the solver is valuable and LLMs are able to learn from it.
To this end, we first modify the source code of the \texttt{pyke}\footnote{\url{https://pyke.sourceforge.net/}} to achieve the following functionalities:
\begin{enumerate}
    \item For each application of a Rule, explicitly state the Rule being \sethlcolor{lightyellow}\hl{\texttt{`Used'}}, or \sethlcolor{lightcyan}\hl{\texttt{`Reused'}} if the Rule has been applied before.
    \item When finishing the application of a Rule, explicitly state the \sethlcolor{lightgreen}\hl{\texttt{`Finish'}} action.
    \item When assigning a value (e.g., an entity) to a variable (e.g., \$x) within a Fact in a Rule, explicitly specify the variable being assigned using \sethlcolor{lightpink}\hl{\texttt{`Bind'}} and its corresponding value.
    \item Similarly, when the variable assignment is complete, provide an explicit indication via \sethlcolor{lightlavender}\hl{\texttt{`Unbind'}}.
    \item When obtaining a new implied Fact, explicitly state the \sethlcolor{lightpeach}\hl{\texttt{`New Fact obtained'}}. If this Fact is an  \sethlcolor{lightaqua}\hl{\texttt{`Already known or implied Fact'}}, this should also be noted explicitly.
    \item Upon the completion of reasoning, explicitly display \sethlcolor{lime}\hl{\texttt{`All newly implied Facts'}} in the knowledge base.
\end{enumerate}

With the aforementioned instructions, we can obtain the revealed solver's reasoning process for the construction of training data. We also highlighted the initial occurrences of each functionality using the corresponding colors in Figure~\ref{fig:ExampleSimplfied} (Turn-2), where a case will be described in detail in the next section.

\subsection{Constructing the Instruction-tuning Data}
\label{sec:construct}

However, as previously mentioned, we cannot guarantee that LMs can definitely complete the NL-to-SL parsing on arbitrary questions. To this end, we first filter out all unsuccessfully parsed training cases that cannot be executed by \texttt{pyke}. Then we reorganize and refine the filtered training data to enhance the interpretability of the solver-derived reasoning steps. For each case, we divide the reasoning process into four conversational turns (Turn-1\&3 for human and Turn-2\&4 for LM), which will be described elaborately in the following paragraphs. We also provide a comprehensive training example of our instruction-tuning data\footnote{In the original case, the Query is `Charlie is not green.'. We replace it with `Dave is not green.' for better illustration.} in Figure~\ref{fig:ExampleSimplfied}, and the full version is also included in Appendix~\ref{apd:instruct_case}.

\paragraph{Turn-1: Instructions \& NL logical Context.} 
For each NL logical question within the training set, we begin by stripping away the specific Query statement while retaining the question Context and subsequently integrating it with elaborately crafted instructions. Taking the case in Figure~\ref{fig:ExampleSimplfied} as an example, we temporarily exclude the Query `Dave is not green' from the `Context' field.
Here, we only consider Query-agnostic question description to ensure that LMs initially focus on the logical background itself. This is because sometimes the ground-truth answer is `Unknown' (e.g., cases in ProofWriter). The truth value of the Query cannot be inferred from the Context, and therefore we need to deduce all implied Facts first.

\paragraph{Turn-2: Query-agnostic Solver-derived Reasoning.}
As we have acquired the solver's symbolic reasoning data in the revealing phase, our goal in Turn-2 is to further refine and enhance the reasoning process to achieve a more readable form of the solver's reasoning process. Specifically, for each logical question, we first define all necessary predicates and append the corresponding natural language explanations. Then we list the known Facts and Rules extracted from the Context with interleaved NL instructions. 

After that, we represent the application of each Rule by utilizing separate blocks, line by line. We strive to preserve as many solver actions as possible, such as `Binding' and `Unbinding', as well as the acquisition of new implied Facts, and so forth. Noting that this information has already been obtained during the revealing phase, we focus on the refinement of the solver-derived reasoning process. Finally, we enumerate all newly implied Facts to enable the model to perform an interim review.

\paragraph{Turn-3: Query \& Answering Instructions.}
In Turn-3, we present instructions for answering a given Query. Following prior works~\citep{ceri1989you,tafjord-etal-2021-proofwriter}, a Query can be considered true within a certain logical context if it is explicitly mentioned or if it can be implied through several Rule applications. To handle negation, we consider two distinct assumptions: 1) the open-world assumption (OWA) that treats any fact that cannot be provable as special truth value `unknown'; 2) the closed-world assumption (CWA) where any fact not provable is assumed `false'. Following both assumptions, we adjust the answering instructions, particularly the `Options' part.

\paragraph{Turn-4: Query-based Reasoning \& Formatted Answer.}
In the final Turn-4, we compare the parsed Query with all the known Facts and implied Facts, expecting the model to perform basic language inference and generate answer options in the desired format.

\subsection{Fine-tuning Open-source LLMs}
\label{sec:finetune}
After obtaining the refined deductive reasoning instruction-tuning dataset, we can perform fine-tuning on open-source LLMs with the expectation that the trained model (i.e., \Model) can possess reasoning abilities similar to those of solvers. Consequently, for any given Query, we can bypass the syntax or grammatical errors derived from NL-to-SL parsing by directly generating the answer with a fine-tuned \Model.
 
\begin{table*}[t!]
\centering
\resizebox{0.85\textwidth}{!}{
    \begin{tabular}{llcc}
    \toprule
        Model & Prompting Methods & ProofWriter & PrOntoQA \\ 
        \midrule
        Random Answering & - & 33.33 & 50.00 \\
        \midrule
        \multicolumn{4}{l}{\textit{\textbf{closed-source LMs}}}               \\
        ChatGPT (\texttt{gpt-3.5-turbo})  & Few-shot Standard & 35.50 &  47.40 \\
        ChatGPT (\texttt{gpt-3.5-turbo})  & Few-shot CoT & 49.17 & 67.80  \\
        GPT-3.5 (\texttt{text-davinci-003})  & Few-shot Standard & 36.16 & 51.80  \\
        GPT-3.5 (\texttt{text-davinci-003})  & Few-shot CoT &  48.33 &  83.00  \\
        GPT-4 (\texttt{gpt-4})  & Few-shot Standard & 52.67  & 77.40  \\
        GPT-4 (\texttt{gpt-4})  & Few-shot CoT & 68.11 & \underline{98.79}  \\
        \midrule
        \multicolumn{4}{l}{\textit{\textbf{open-source LMs}}}               \\
        Vicuna-13B (\texttt{vicuna-13b-v1.5-16k})  & Few-shot Standard & 35.50 & 53.80  \\
        Vicuna-13B (\texttt{vicuna-13b-v1.5-16k}) & Few-shot CoT & 41.50 &  37.40 \\
        CodeLlama-13B-Base (\texttt{CodeLlama-13b-hf}) & Few-shot Standard & 0.00 & 0.00  \\
        CodeLlama-13B-Base (\texttt{CodeLlama-13b-hf}) & Few-shot CoT & 36.00 & 50.00  \\
        CodeLlama-13B-Instruct (\texttt{CodeLlama-13b-Instruct-hf}) & Few-shot Standard & 36.83 & 52.20   \\
        CodeLlama-13B-Instruct (\texttt{CodeLlama-13b-Instruct-hf}) & Few-shot CoT & 32.67 & 66.40  \\
        \midrule
        \multicolumn{4}{l}{\textit{\textbf{solver-argumented LMs}}}               \\
        LogicLM (\texttt{gpt-3.5-turbo}) & Few-shot CoT & 58.33 & 61.00 \\
        LogicLM (\texttt{text-davinci-003}) & Few-shot CoT & 71.45 & 85.00 \\ 
        LogicLM (\texttt{gpt-4}) & Few-shot CoT & 79.66 & 83.20 \\

        \midrule
        \multicolumn{4}{l}{\textit{\textbf{ours}}}               \\
        \Model~(\texttt{vicuna-13b-v1.5-16k}) & Four-turn CoT & 81.17 & \textbf{96.40}  \\
        \Model~(\texttt{CodeLlama-13b-hf}) & Four-turn CoT & \underline{\textbf{89.50}} & 95.60 \\
        \Model~(\texttt{CodeLlama-13b-Instruct-hf}) & Four-turn CoT & 81.67 & 96.20  \\

\bottomrule
   
    \end{tabular}
} 
    \caption{Main results on two evaluation datasets. The best results of \Model~are in \textbf{bold} and the best results within each dataset are \underline{underlined}.
    }
    \label{tbl:main_results}
\end{table*}

\section{Experiments}
We construct our solver-derived instruction-tuning data on two public deductive reasoning datasets and evaluate \Model~on corresponding test sets.

\subsection{Datasets}
\paragraph{ProofWriter}\citep{tafjord-etal-2021-proofwriter} is a commonly employed dataset for deductive logical reasoning. Following~\citet{pan2023logic}, we adopt the open-world assumption (OWA) subset where the answer of each example is one of \{\textit{True}, \textit{False}, \textit{Unknown}\}. The original dataset is partitioned into 5 subsets where each part requiring 0, $\le$1, $\le$2, $\le$3, and $\le$5 hops of reasoning, respectively. For evaluation, we adopted the version provided by~\citet{pan2023logic}, which comprises 600 samples from the most challenging 5-hop subsets with balanced label distribution. For training, we merged all training subsets and obtained 41,433 training examples after the construction stage.

\paragraph{PrOntoQA}\citep{saparov2023language} is a synthetic logical reasoning dataset created recently to test the general deductive reasoning capacity of LLMs. We adopt the hardest \textit{fictional characters} version of the dataset following~\citet{pan2023logic} where the entities of Facts are fictional concept names (e.g., `wumpus' instead of `cat'), to avoid any confounding effects from knowledge acquired during the pretraining phase. Similar to ProofWriter, PrOntoQA is organized into several subsets based on the number of required reasoning steps. We use the hardest 5-hop subset for evaluation. Contrary to ProofWriter, PrOntoQA is in a closed-world assumption (CWA) subset where the answer of each example is one of \{\textit{True}, \textit{False}\}. For training, we merely merge all subsets with \textit{fictional characters} and obtained 15,940 training cases after filtering out syntax-invalid ones.

\subsection{Baselines}
We consider comparing \Model~with following groups of baselines: 
\paragraph{Closed-source LMs:} We include the ChatGPT (\texttt{gpt-3.5-turbo}) \citep{chatgpt2022}, GPT-3.5 (\texttt{text-davinci-003}) \citep{ouyang2022training} and GPT-4 (\texttt{gpt-4}) \citep{openai2023gpt4} as closed-source LMs for evaluation following~\citet{pan2023logic}.

\paragraph{Open-source LMs:} We also evaluate open-source LMs for research community. Specifically, we choose Vicuna-13B (\texttt{vicuna-13b-v1.5-16k}) \citep{vicuna2023}, a chatbot trained by fine-tuning LLaMA-2 \citep{touvron2023llama2} on user-shared conversations collected from ShareGPT\footnote{\url{https://sharegpt.com/}}, and CodeLlama-13B \citep{roziere2023code}, foundation models for code tasks. We select the base version (\texttt{CodeLlama-13b-hf}), and instruction fine-tuned version (\texttt{CodeLlama-13b-Instruct-hf}).

\paragraph{Solver-argumented LMs:} Finally, we compare our model against the solver-argumented LMs. We focus on the representative LogicLM~\citep{pan2023logic} with underlying LLMs ChatGPT (\texttt{gpt-3.5-turbo}), GPT-3.5 (\texttt{text-davinci-003}) and GPT-4 (\texttt{gpt-4}), which serve as the state-of-the-art deductive reasoning methods. 

Apart from the LMs, we also analyze two types of prompting methods: i) \textit{Standard prompting} that uses in-context learning with few-shot demonstrations to directly answer the given question; ii) \textit{Chain-of-Thought (CoT)} that utilizes step-by-step problem-solving process to generate explanations where few-shot demonstrations are also provided, and then outputs the final answer. For a fair comparison, we use the same in-context examples, shown in Appendix~\ref{apd:instructions_proofwriter} and~\ref{apd:instructions_prontoqa}, for NL-to-SL parsing when evaluating all models on the same dataset, consistent with~\citet{pan2023logic}. To enhance the clarification, we also provide a specific baseline \textit{`Random Answering'} that randomly outputs answer options.

\subsection{Implementation Details}

During the fine-tuning phase, we use a batch size of 32 per GPU and a learning rate of 1e-5 for all open-source LMs. We train our model on 8 Nvidia A100-80G GPUs with DeepSpeed ZeRO-3~\citep{rasley2020deepspeed} for 12 hours on 2 epochs. For reproducibility, we use greedy decoding and set the temperature to 0 and the maximum context length to 8192. As for baselines, we strictly follow the setting of ~\citet{pan2023logic}. Given that all instances are presented in the form of multiple-choice questions, we assess the model's performance by the accuracy of selecting the correct answer option.

\subsection{Main Results}
We report the results of \Model~and baselines on Table~\ref{tbl:main_results} and have following main findings:

1) \ul{When prompting with few-shot examples, open-source LMs exhibit notably poor deductive reasoning capabilities, with their outputs closed to random answering.} Even the Standard prompting models of ChatGPT (\texttt{gpt-3.5-turbo}) and GPT-3.5 (\texttt{text-davinci-003}) exhibit a similar performance to random answering. This once again demonstrates that it is considerably difficult for many LLMs to solve logical reasoning tasks.

2) \ul{\Model~is significantly superior to the state-of-the-art solver-augmented LMs by a large margin on both deductive reasoning benchmarks.} In ProofWriter, our best-performing model, \Model~(\texttt{CodeLlama-13b-hf}), outperforms the currently state-of-the-art LogicLM (\texttt{gpt-4}) by an absolute improvement of 9.84\%. Meanwhile, in PrOntoQA, our best-performing model \Model~(\texttt{vicuna-13b-v1.5-16k}) exhibits an even higher absolute improvement of 13.20\% than LogicLM (\texttt{gpt-4}). This indicates that our approach is better than the pipeline of problem formulation first and then reasoning with solvers, and fine-tuning with solver-derived reasoning data can facilitate the deductive reasoning capacity of LMs.

3) \ul{\Model~significantly outperforms all selected open/closed-source LMs on both datasets, except for the CoT experiment on the PrOntoQA data where \Model~achieves comparable results with GPT-4 CoT.} This is surprising considering that our underlying open-source LMs are merely 13B parameters in size. As for the baseline experiments of GPT-4, our performance on ProofWriter also significantly surpasses that of GPT-4's Standard and CoT prompting versions, as well as the Standard version of PrOntoQA. These results further demonstrate that open-source LMs, when coupled with solver-simulated reasoning capacity, can achieve performance on par with or even superior to closed-source GPT models.

4) \ul{The accuracy of CodeLlama-13B-Base (\texttt{CodeLlama-13b-hf}) with Standard prompting was 0.00, and the performance of the CoT version was close to random answering.} By examining the outputs, we found that this is due to the CodeLlama-13B-Base's inability to follow the provided few-shot demonstrations, resulting in outputting no answering options. The introduction of the Instruct version of CodeLlama-13B mitigates this issue to some extent. However, after training with \Model, the CodeLlama models far less encounter this issue (i.e., following the right answering format in both test sets) and even achieve better performance than the Vicuna version of \Model. This demonstrates the potential of code foundation models in logical reasoning tasks, consistent with the finding on prior work~\citep{yue2023mammoth}.

\begin{table}[!t]
\centering
\resizebox{0.45\textwidth}{!}{
    \begin{tabular}{lc}
     \toprule
        Model  & Accuracy \\ 
        \midrule
        \Model~(\texttt{vicuna-13b-v1.5-16k})  & 81.17 \\
        \midrule
        + (w/o \texttt{`unbind'} statements)  & 80.67 \\
        + (w/o \texttt{`fail \& backtrack'} statements)  & \underline{84.00} \\
        + (w/ NL representation)  & 66.33   \\
        \midrule
        \midrule
        \Model~(\texttt{CodeLlama-13b-hf})  & 89.50 \\
        \midrule
        + (w/o \texttt{`unbind'} statements)  & \underline{93.33} \\
        + (w/o \texttt{`fail \& backtrack'} statements) & 87.17 \\
        + (w/ NL representation) & 52.33  \\
        \midrule
        \midrule
        \Model~(\texttt{CodeLlama-13b-Instruct-hf})  & 81.67 \\
        \midrule
        + (w/o \texttt{`unbind'} statements) & 79.00 \\
        + (w/o \texttt{`fail \& backtrack'} statements) & \underline{84.83} \\
        + (w/ NL representation)  & 66.33  \\

\bottomrule
    
    \end{tabular}
} 
    \caption{The accuracy of the variations on solver-derived reasoning format, and replacing SL representations with NL on ProofWriter. The best results on each underlying LMs are \underline{underlined}.}
    \label{tbl:variations_result}
\end{table}

\begin{table}[!t]
\centering
\resizebox{0.48\textwidth}{!}{
    \begin{tabular}{llccc}
     \toprule
        Train set & Test Set & VCN & CLB & CLI \\ 
        \midrule
         PrOntoQA & PrOntoQA & 96.40 & 95.60 & 96.20 \\
         \midrule
         Both  & PrOntoQA  & 91.00 & 87.00 & 89.00 \\ 
         Both (Reformat) & PrOntoQA  & 90.00 & 87.00  & 77.80 \\ 
        \midrule
        \midrule
         ProofWriter & ProofWriter & 81.17  & 89.50 & 81.67  \\
         \midrule
         Both  & ProofWriter  & 79.33  & 87.17   & 79.67 \\ 
         Both (Reformat) & ProofWriter  & 79.00 & 90.83  & 84.50  \\ 

        \bottomrule
    \end{tabular}
} 
    \caption{The accuracy of \Model~trained with merged data and tested on single data with different underlying LMs. `VCN', `CLB', and `CLI' respectively represent Vicuna-13B, CodeLlama-13B-Base, and CodeLlama-13B-Instruct. `Both' means `ProofWriter + PrOntoQA'.} 
    \label{tbl:transfer_result}
\end{table}

\section{Further Analysis}

\subsection{Impact of Solver-derived Reasoning Formats}

We further investigate the impact of different solver-derived reasoning formats on the model's performance. Specifically, we consider the following format variations: 1) \textit{w/o `unbind' statements} that we remove all \sethlcolor{lightlavender}\hl{\texttt{`Unbind'}} statements from \textbf{Turn-2} to investigate the utility of the explicit retention of this action from the solver; 2) \textit{w/o `fail \& backtrack' statements} that we removing all \sethlcolor{lightgray}\hl{\texttt{`Fail \& backtrack'}} statements from \textbf{Turn-2}. During the solver's reasoning process, it is expected to encounter situations in which, after binding a value, the solver realizes that not all premises are satisfied (e.g., \texttt{`Fiona is blue'} but \texttt{`Fiona is not quiet'} for application of Rule3 in Figure~\ref{fig:ExampleSimplfied}). Consequently, a \sethlcolor{lightgray}\hl{\texttt{`Fail \& backtrack'}} operation occurs (highlighted in color in Figure~\ref{fig:ExampleSimplfied}). We explore the effectiveness of explicitly stating these operations. 

We present the accuracy of the variations on solver-derived reasoning format on ProofWriter in Table~\ref{tbl:variations_result} where several observations can be made: 1) regardless of using the default format, removing \texttt{`Unbind'} statements, or removing \texttt{`Fail \& backtrack'} statements, it can not be determined which format guarantees the optimal results. To retain the maximum amount of action information that the solver can provide, we still adopt the default settings in \Model; 2) whether \texttt{`Unbind'} statements are removed or \texttt{`Fail \& backtrack'} statements are removed, there is always an experiment under each open-source LMs that can surpass the default \Model~results. This further enhances the best performance of \Model~shown in Table~\ref{tbl:main_results}.

\subsection{Impact of SL Reasoning Representations}

We are also curious about the impact of SL reasoning representations. Therefore, we include additional experiments in Table~\ref{tbl:variations_result}, denoted as \textit{w/ NL representation} that we re-translate the symbolic representation (e.g., \texttt{Green(’Charlie’, True)}) back to its original NL version (e.g., \texttt{Charlie is green.}) and replace the original symbolic representation in \textbf{Turn-2}. From the table, we can find that replacing SL representations with NL results in a significant decrease in model performance, further emphasizing that symbolic representations are superior to NL representations in deductive reasoning tasks.

\subsection{Effectiveness of Merging Data from Different Reasoning Assumptions}

Since ProofWriter is an open-world assumption and PrOntoQA is labeled within a closed-world assumption, we also perform a further investigation on whether both reasoning assumptions can benefit each other. Specifically, we first merge both constructed training data and then test \Model~on each test set. The experimental results are shown in Table~\ref{tbl:transfer_result}. We can conclude that if we directly mix the two types of data for training, the results on their respective test sets will be slightly lower than those obtained from training solely on their respective datasets. Therefore, we conducted an in-depth analysis of the underlying reasons and observed that in PrOntoQA, the majority of Rules are in the format of `Every/Each \texttt{A} is (not) \texttt{B}' or `\texttt{A} are (not) \texttt{B}'. While in ProofWriter, the predominant structure of Rules consists of: `If someone is \texttt{A}, then they are \texttt{B}' or `If something is \texttt{A}, then it is \texttt{B}'. Therefore, we conducted an additional set of experiments in which the Rule format of two training sets was randomly reformatted into the four aforementioned types using regular expression (denoted as `Both (Reformat)'). Then, we test the model on the original test sets. We can observe that by employing this approach, the code models yield improved performance on ProofWriter. Thus, the style/genre of logical context must also be taken into consideration to maximize the efficacy of transfer learning in logical reasoning.

\section{Related Work}

\paragraph{Logical Reasoning with LMs.} Recent efforts in adapting Large Language Models (LLMs) for logical reasoning tasks generally adopt direct fine-tuning specialized modules~\citep{clark2020transformers,tafjord-etal-2021-proofwriter,tafjord-etal-2022-entailer,yang-etal-2022-generating} or in-context learning~\citep{zhou2022least,lyu2023faithful,ling2023deductive}, where reasoning in NL is used by both groups of methods. Fine-tuning approaches involve training the full model or specialized modules, enhancing LLMs with module-level logical reasoning skills like proof, enumeration, and abduction~\citep{tafjord-etal-2021-proofwriter}. The in-context learning approaches create specific prompts to encourage LLMs' step-by-step reasoning skills. Common methods encompass chain-of-thought prompting~\citep{wei2022chain, chen2023chatcot}, which produces explanations before delivering a final answer, and least-to-most prompting~\citep{zhou2022least}, which deconstructs a problem into simpler components that can be resolved individually. Some recent work has focused on combining neural networks with symbolic reasoning~\citep{tian2022weakly, pryor2022neupsl, pan2023logic}, especially the solver-augmented LMs that parse NL logical questions into symbolic representations, then utilizing external logical solvers for answering. Despite their impressive performance, parsing errors can lead to solver execution failure and logical question-answering issues. To address this, we propose \Model, which directly imitates the solver's reasoning ability and outputs the answer.

\paragraph{Augmented LMs for Reasoning.} Recent work has begun to augment LMs to overcome their inherent limitations such as the incapacity to access up-to-date information or conduct accurate mathematical reasoning. They augment with external tools and resources, such as the information retriever~\citep{shi2023replug, lazaridou2022internet}, planner~\citep{liu2023llm+} and other pre-trained models~\citep{shen2023hugginggpt}. Specifically, to enhance the reasoning capacity, recent work resort to external off-the-shelf \textbf{Solvers} including programmatic interpreters~\citep{chen2022program,gao2023pal}, satisfiability solvers~\citep{ye2023satisfiability}, logical solvers~\citep{pan2023logic} or their hybrids~\citep{poesia2023certified}. Most of them utilize the LMs to parse the NL question to symbolic representations and then invoke solvers to reason in SL. In this paper, we concentrate on logical solvers, automated tools for validating the truth value of logical formulas.
\section{Conclusion}
In this paper, we propose a novel \Model~that can directly act as a logical solver for deductive reasoning tasks. \Model~can output all facts implied from NL logical questions, while bypassing the syntax or grammatical errors derived from NL-to-SL parsing of solver-augmented LMs. We conducted numerous analytical experiments on two public deductive reasoning benchmarks. Evaluation results show that \Model~can significantly outperform state-of-the-art solver-augmented LMs, and surpass or be comparable with few-shot prompting methods on competitive LLMs like ChatGPT or GPT-4.

\bibliography{custom}

\newpage

\appendix

\newpage

\section{Instructions for NL-to-SL Parsing on ProofWriter}
\label{apd:instructions_proofwriter}

\begin{caseBoxEnv}{} \small
\textbf{Task Description:} You are given a problem description and a question. The task is to:\\
1) define all the predicates in the problem\\
2) parse the problem into logic rules based on the defined predicates\\
3) write all the facts mentioned in the problem\\
4) parse the question into the logic form\\
\\
\textbf{Problem:}\\
Anne is quiet. Erin is furry. \textit{(... more context here ...)} All red people are young.\\
\\
\textbf{Question:}\\
Based on the above information, is the following statement true, false, or unknown? Anne is white.\\
\\
\textbf{Predicates:}\\
Quiet(\$x, bool) ::: Is x quiet?\\
Furry(\$x, bool) ::: Is x furry?\\
\textit{(... more predicates here ...)} \\
Young(\$x, bool) ::: Is x young?\\
\\
\textbf{Facts:}\\
Quite(Anne, True) ::: Anne is quiet.\\
Furry(Erin, True) ::: Erin is furry.\\
\textit{(... more facts here ...)} \\
Quite(Harry, True) ::: Harry is quiet.\\
White(Harry, True) ::: Harry is white.\\
\\
\textbf{Rules:}\\
Young(\$x, True) $>>>$ Furry(\$x, True) ::: Young people are furry.\\
Quite(Anne, True) $>>>$ Red(\$x, True) ::: If Anne is quiet then Anne is red.\\
\textit{(... more rules here ...)} \\
Red(\$x, True) $>>>$ Young(\$x, True) ::: All red people are young.\\
\\
\textbf{Query:}\\
White(Anne, True) ::: Anne is white.\\
------\\
\textbf{Problem:}\\
\textit{(new problem here)}\\
\textbf{Question:}\\
\textit{(new question here)}\\  
\end{caseBoxEnv}

\newpage

\section{Instructions for NL-to-SL Parsing on PrOntoQA}
\label{apd:instructions_prontoqa}

\begin{caseBoxEnv}{} \small
\textbf{Task Description:} You are given a problem description and a question. The task is to:\\
1) define all the predicates in the problem\\
2) parse the problem into logic rules based on the defined predicates\\
3) write all the facts mentioned in the problem\\
4) parse the question into the logic form\\
\\
\textbf{Problem:}\\
Each jompus is fruity. Every jompus is a wumpus. \textit{(... more context here ...)} Alex is a tumpus.\\
\\
\textbf{Question:}\\
True or false: Alex is not shy.\\
\\
\textbf{Predicates:}\\
Jompus(\$x, bool) ::: Does x belong to Jompuses?\\
Fruity(\$x, bool) ::: Is x fruity?\\
\textit{(... more predicates here ...)} \\
Liquid(\$x, bool) ::: Is x liquid?\\
Zumpus(\$x, bool) ::: Does x belong to Zumpuses?\\
\\
\textbf{Facts:}\\
Tumpus(Alex, True) ::: Alex is a tumpus.\\
\\
\textbf{Rules:}\\
Jompus(\$x, True) $>>>$ Fruity(\$x, True) ::: Each jompus is fruity.\\
Jompus(\$x, True) $>>>$ Wumpus(\$x, True) ::: Every jompus is a wumpus.\\
\textit{(... more rules here ...)} \\
Rompus(\$x, True) $>>>$ Liquid(\$x, True) ::: Each rompus is liquid.\\
Rompus(\$x, True) $>>>$ Zumpus(\$x, True) ::: Rompuses are zumpuses.\\
\\
\textbf{Query:}\\
Shy(Alex, False) ::: Alex is not shy.\\
------\\
\textbf{Problem:}\\
\textit{(new problem here)}\\
\textbf{Question:}\\
\textit{(new question here)}\\
\end{caseBoxEnv}

\section{A Comprehensive Training Example of our Instruction-tuning Data}
\label{apd:instruct_case}

\begin{figure*}[t!]
  \centering   
  \includegraphics[width=0.95\textwidth]{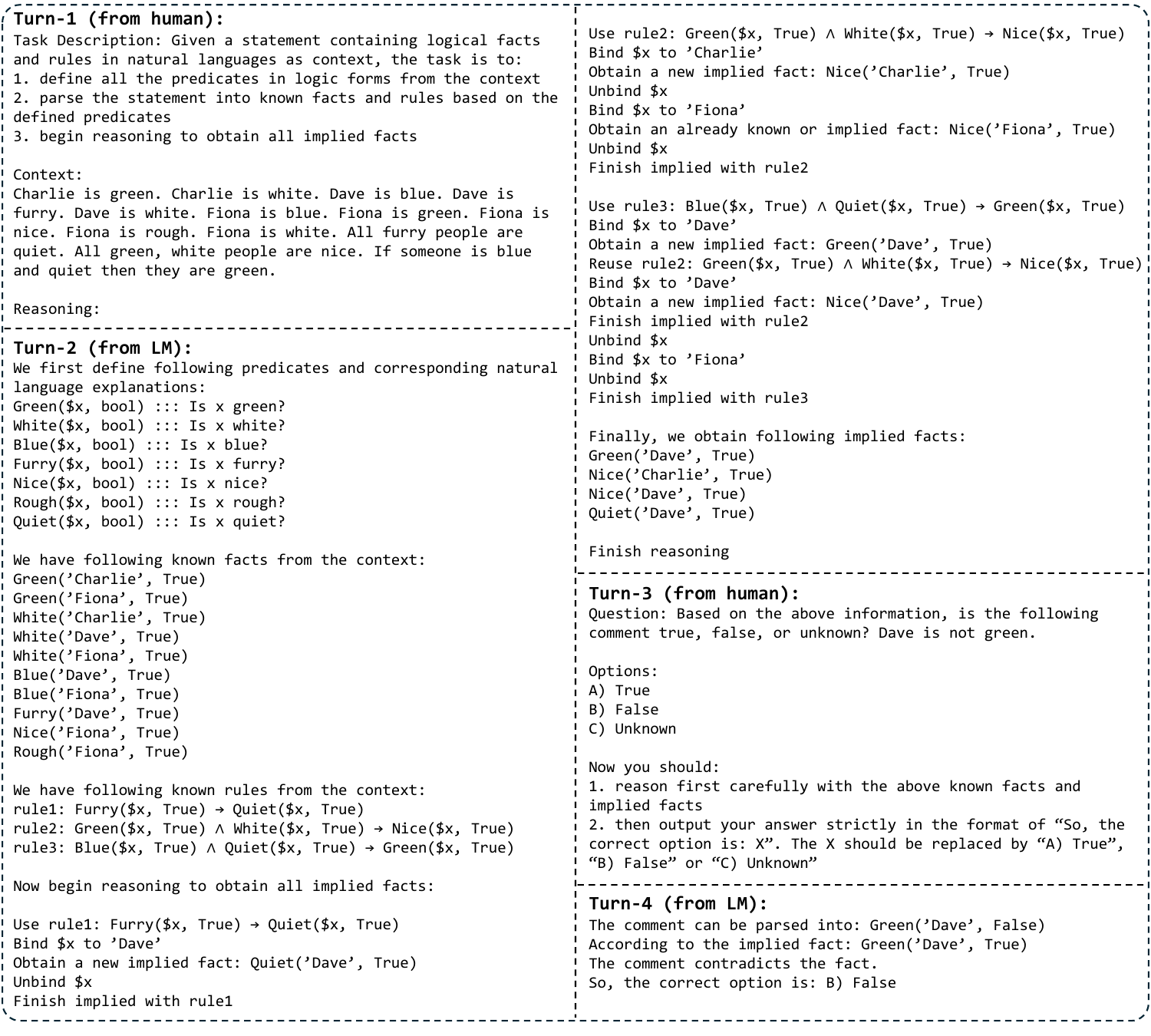}
  \caption{The full version of the comprehensive 4-turn training example of our instruction-tuning data shown in Figure~\ref{fig:ExampleSimplfied}.}
 \label{fig:Example}
\end{figure*}

\end{document}